\documentclass[runningheads]{llncs}
\usepackage[T1]{fontenc}
\usepackage{graphicx}
\usepackage{hyperref}
\usepackage{color}

\usepackage{multirow}
\usepackage{tabularx}
\usepackage{booktabs}

\begin{document}
%

\title{Towards Abdominal 3-D Scene Rendering from Laparoscopy Surgical Videos using NeRFs}

\titlerunning{Laparoscopic NeRF}
%

\author{Khoa Tuan Nguyen\inst{1,2} \and
Francesca Tozzi\inst{4,6} \and
Nikdokht Rashidian\inst{5,6} \and \\
Wouter Willaert\inst{4,6} \and
Joris Vankerschaver\inst{2,3} \and
Wesley De Neve\inst{1,2}}


%
\authorrunning{Khoa et al.}
%


\institute{
IDLab, ELIS, Ghent University, Ghent, Belgium \and
Center for Biosystems and Biotech Data Science, Ghent University Global Campus, Incheon, Korea \and
Department of Applied Mathematics, Informatics, and Statistics, Ghent University, Ghent Belgium \\
\email{\{khoatuan.nguyen,joris.vankerschaver,wesley.deneve\}@ghent.ac.kr} \and 
Department of GI Surgery, Ghent University Hospital, Ghent, Belgium \and
Department of HPB Surgery \& Liver Transplantation, Ghent University Hospital, Ghent, Belgium \and
Department of Human Structure and Repair, Ghent University, Ghent, Belgium \\
\email{\{francesca.tozzi,nikdokht.rashidian,wouter.willaert\}@ugent.be}
}


%
\maketitle              
\begin{abstract}
Given that a conventional laparoscope only provides a two-dimensional (2-D) view, the detection and diagnosis of medical ailments can be challenging.
To overcome the visual constraints associated with laparoscopy, the use of laparoscopic images and videos to reconstruct the three-dimensional (3-D) anatomical structure of the abdomen has proven to be a promising approach.
Neural Radiance Fields (NeRFs) have recently gained attention thanks to their ability to generate photorealistic images from a 3-D static scene, thus facilitating a more comprehensive exploration of the abdomen through the synthesis of new views.
This distinguishes NeRFs from alternative methods such as Simultaneous Localization and Mapping (SLAM) and depth estimation.
In this paper, we present a comprehensive examination of NeRFs in the context of laparoscopy surgical videos, with the goal of rendering abdominal scenes in 3-D.
Although our experimental results are promising, the proposed approach encounters substantial challenges, which require further exploration in future research.
 
\end{abstract}
%
%
%

\section{Introduction}

\begin{figure}[tb]
    \centering
    \includegraphics[width=1\textwidth]{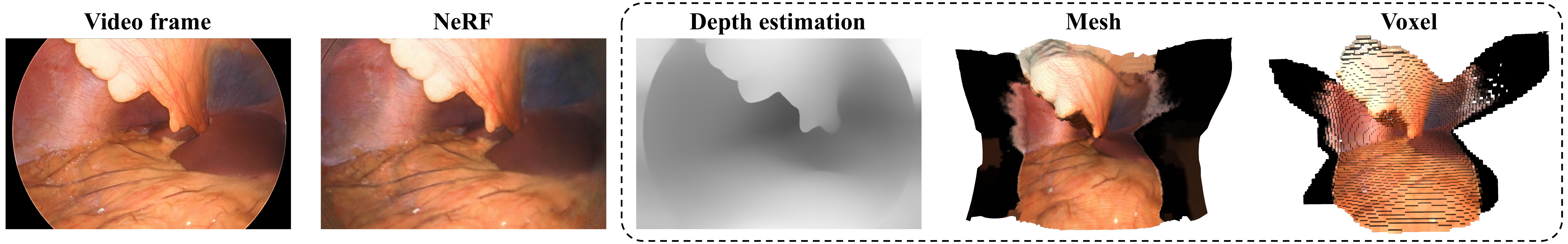}
    \caption{
    Illustration of NeRFs and 3-D reconstruction using estimated depth information.
    From left to right: a video frame extracted from a laparoscopic surgical video, an image rendered by NeRFs, depth estimation for the given video frame, and reconstructions in both mesh and voxel formats based on the estimated depth information.
    }
    \label{fig:abstract_demo}
\end{figure}

Laparoscopy, also known as keyhole surgery or minimally invasive surgery (MIS), is a surgical technique that enables a surgeon to access the inside of the abdomen without the need for making large incisions in the skin.
The surgeon inserts a slender tool, with a light and a camera attached, through small skin incisions, which makes it then possible for the surgeon to see inside the abdomen, negating the need for large skin cuts and a long recovery period.
However, the view of the surgeon during MIS is limited to the perspective of the camera. 
To address this limitation, various approaches have been developed to provide surgeons with additional information such as depth, segmentation, 3-D reconstruction, and augmented surgery techniques
~\cite{ali2022we,lin2016video,soler2021augmented}.
Despite substantial advancements, current methods still face challenges in achieving high-quality and photorealistic 3-D reconstruction results, as shown in Fig.~\ref{fig:abstract_demo}.
Neural Radiance Fields (NeRFs)~\cite{mildenhall2020nerf,mueller2022instant} have recently emerged as a popular method for view synthesis using natural images. However, their application in the medical field has thus far remained limited~\cite{gerats2022depth,9761457,wang2022neural}. 
In this paper, we focus on the task of understanding the underlying scene in MIS through the utilization of NeRFs. 
Specifically, we investigate how well NeRFs can be applied to laparoscopic surgical videos, particularly during the preparatory phase of a surgical procedure where surgical tools, surgical actions, scenes with blood, and moving organs are absent.
In summary, our study makes the following contributions:

\begin{itemize}
    \item We present a workflow that analyzes laparoscopic surgical videos by employing state-of-the-art NeRFs-based methods to render a complete abdominal environment.

    \item We adopt mask-based ray sampling to better mitigate the presence of artefacts that are associated with the black outer areas commonly found in laparoscopic surgical videos.
    
    \item We investigate the performance of each adopted NeRFs-based method, identifying and discussing the challenges associated with their usage.
\end{itemize}

\section{Related Work}

One approach to achieve 3-D scene reconstruction is through the utilization of Simultaneous Localization And Mapping (SLAM), which enables the mapping of 3-D coordinates and the direct reconstruction of 3-D surfaces from a given camera view~\cite{lin2016video}.
Guodong~\textit{et al.} proposed an enhanced SLAM method called MKCF-SLAM~\cite{wei2020novel,9588016}, incorporating multi-scale feature patch tracking to reconstruct laparoscopic scenes characterized by texture deficiency, high levels of reflection, and changes in target scale.
In contrast, Haoyin~\textit{et al.} presented a 2-D non-rigid SLAM system known as EMDQ-SLAM~\cite{zhou2021emdq,9374480}, capable of compensating for pixel deformation and performing real-time image mosaicking.
Another approach for 3-D scene reconstruction involves depth estimation. 
Recasen~\textit{et al.} proposed Endo-Depth-and-Motion~\cite{recasens2021endo}, a pipeline for 3-D scene reconstruction from monocular videos that combines self-supervised depth estimation, photometric odometry, and volumetric fusion. 
Additionally, Baoru~\textit{et al.}~\cite{huang2022self} leveraged stereo image pairs to generate corresponding point cloud scenes and minimize the point distance between the two scenes.
Shuwei~\textit{et al.}~\cite{shao2022self} proposed a fusion of depth prediction, appearance flow prediction, and image reconstruction to achieve robust training.

Unlike the methods discussed previously, NeRFs are based on deep neural networks that focus on learning to represent 3-D scenes, as proposed by Mildenhall~\textit{et al.}~\cite{mildenhall2020nerf}. 
To the best of our knowledge, the application of NeRFs to surgical videos, and endoscopic videos in particular, was first discussed in the work of Wang~\textit{et al.}, resulting in an approach known as EndoNeRF~\cite{wang2022neural}.
EndoNeRF specifically operates on prostatectomy data and leverages the D-NeRF method~\cite{pumarola2021d} to render a short dynamic scene and inpaint the surgical tools.
However, whereas the focus of EndoNeRF is on rendering a specific area, the focus of our research effort is on rendering an entire abdominal environment.
Moreover, our research effort primarily focuses on rendering static scenes, with a minimal presence of surgical tools, surgical actions, and moving organs. 
In particular, we generate static scenes through the use of two NeRFs-based methods that offer fast convergence: NerfAcc~\cite{li2023nerfacc,nerfstudio} and Instant-NGP~\cite{mueller2022instant}.

\section{Method}
\begin{figure}[htb]
    \centering
    \includegraphics[width=1\textwidth]{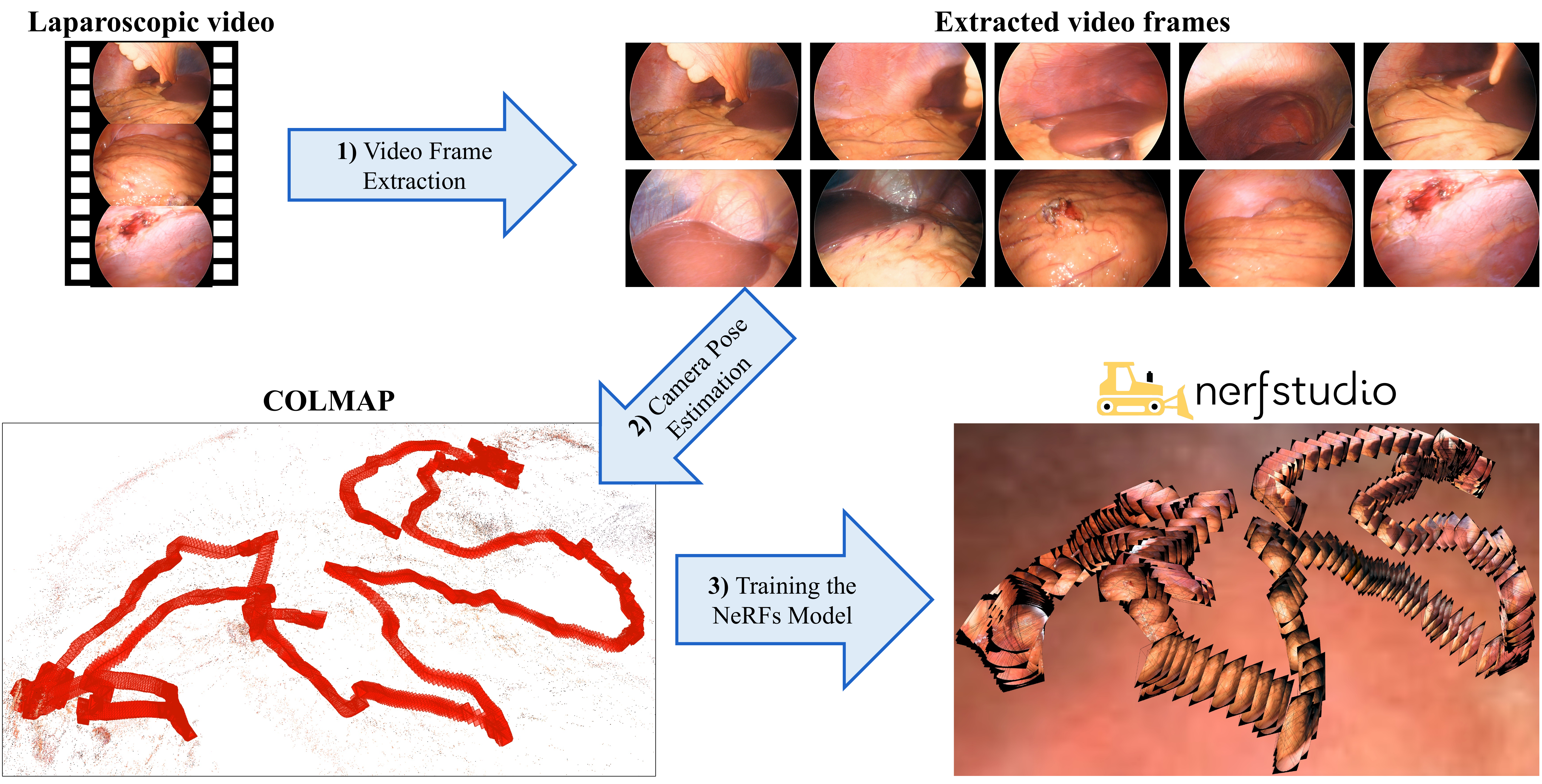}
    \caption{
    NeRFs-based workflow for rendering of the abdomen from a laparoscopic surgical video. 
    The workflow consists of three major steps, including two pre-processing steps and a training step.
    First, a dataset is created by extracting a total of $n_{frame}$ video frames from the laparoscopic surgical video.
    Second, we estimate the camera pose for each extracted video frame, resulting in $n_{pose}$ camera poses.
    The figure displays $n_{pose}$ camera poses obtained using COLMAP after the second step.
    Finally, NeRFs-based models are trained using the extracted images and their corresponding camera poses, utilizing the Nerfstudio library.
    }
    \label{fig:workflow}
\end{figure}

As depicted in Fig.~\ref{fig:workflow}, we present a workflow to apply NeRFs to laparoscopic surgical videos. 
We describe each step of this workflow in the following sections.

\subsection{Video Frame Extraction}
In this section, we describe the process of extracting video frames from a selected laparoscopic surgical video to create a dataset.
In particular, we extract a total of $n_{frame}$ equidistant frames from the given video.
These extracted frames are then stored for subsequent analysis.
By varying the value of $n_{frame}$, we generate multiple datasets, each containing a different number of video frames. 
This allows us to investigate the impact of the variable $n_{frame}$ in our experiments, which encompass a broad spectrum of sparsity levels, ranging from sparsely sampled views (with only a few frames) to densely sampled views (with numerous frames).

\subsection{Camera Pose Estimation}
Upon obtaining a dataset consisting of $n_{frame}$ extracted video frames, it becomes necessary to determine the camera pose for each individual frame.
In order to accomplish this task, we employ Structure-from-Motion (SfM) algorithms that make use of COLMAP~\footnote{\url{https://colmap.github.io/}}~\cite{schoenberger2016sfm,schoenberger2016mvs} with Scale-Invariant Feature Transform (SIFT) features, as well as the hierarchical localization toolbox (HLOC)~\footnote{\url{https://github.com/cvg/Hierarchical-Localization}}~\cite{sarlin2019coarse}. 
Specifically, the HLOC toolbox utilizes COLMAP with SuperPoint~\cite{detone2018superpoint} for feature extraction and SuperGlue~\cite{sarlin2020superglue} for image matching. 
In addition, we also make use of LocalRF~\cite{meuleman2023localrf} as an alternative camera pose estimation method.
This method is capable of handling long-range camera trajectories and large unbounded scenes.
By leveraging these tools, from the initial $n_{frame}$ frames, we can estimate camera poses for a total of $n_{pose}$ frames.
Ideally, by using LocalRF~\cite{meuleman2023localrf}, we would have $n_{frame} = n_{pose}$.

\subsection{Training and Mask-based Ray Sampling}
\label{subsec:method_NeRF}
For training the NeRFs-based models, we utilize the Nerfstudio~\cite{nerfstudio} library, which includes the NerfAcc~\cite{li2023nerfacc} and Instant-NGP~\cite{mueller2022instant} models.
\begin{figure}[htb]
    \centering
    \includegraphics[width=1\textwidth]{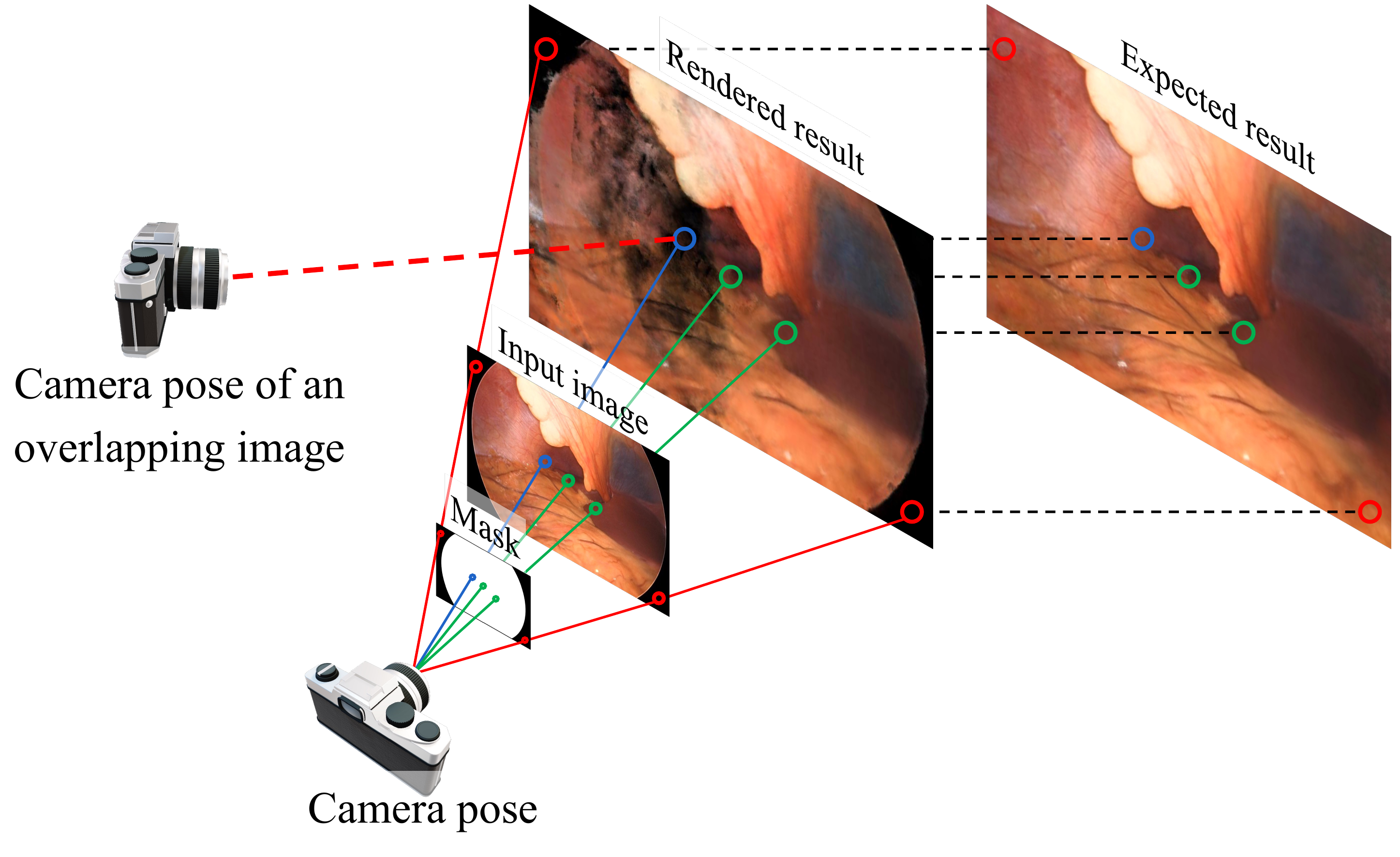}
    \caption{
        Ray sampling for NeRFs training.
        Each line represents a ray.
        Green lines indicate the required sampled rays.
        Red lines represent rays in the black area that should not be sampled.
        Artifacts occur in the rendered result when an image overlaps with the black area of other images. 
        This is demonstrated by the existence of artifacts within the blue circle, which illustrates the point where the blue line intersects with the red lines from other images (represented by the thick dashed red line).
        To obtain the desired result, we create a mask that only samples rays from pixels with color values.
    }
    \label{fig:woMask_problem}
\end{figure}
As depicted in Fig.~\ref{fig:woMask_problem}, our approach involves sampling rays from the input image and its corresponding camera pose, with each ray passing through a pixel on the image. 
It is important to note that we exclude the red lines in order to prevent the occurrence of artifacts at the intersections between these red lines and the blue lines from overlapping images.
To that end, 
we propose sampling rays with masks.
Instead of creating a mask for each individual image, we opted to use a single mask that applies to all images. 
This one-for-all mask takes the form of a circle located at the center of the image, with the radius calculated based on the circular view of the laparoscope.
This mask allows us to selectively sample rays only from pixels that contain color values. 
By employing this mask, the model is able to infer the content of the black areas solely based on the color information from the overlapping regions in other images.

\section{Experiments}
To experiment with the use of NeRFs for rendering an abdominal scene, we chose a laparoscopic surgical video from a collection of videos recorded at Ghent University Hospital.
This chosen video captures the preparatory phase of a laparoscopic surgery and does not involve the use of surgical tools. 
It provides a laparoscopic view recorded during an examination of the abdomen, captured at a resolution of $3840 \times 2160$ pixels.
To optimize the video for our purposes, we cropped out $6.6\%$ from the left and $9.1\%$ from the right, removing redundant black areas.
This resulted in video frames having a resolution of $3237 \times 2160$ pixels.
The recorded footage has a duration of $93$ seconds, and the frame rate is approximately 29.935 fps (frames per second), leading to a total of $2784$ available video frames.
%
All execution times were obtained on a machine 
equipped with a single RTX A6000 GPU and 128 GB RAM.
%


\subsection{Camera Pose Estimation}
We extract video frames from a selected laparoscopic surgical video, with $n_{frame}$ having the following values: 310, 1392, and 2784 frames. 
This results in three datasets, each having a different sampling interval. 
By doing so, we can assess the effectiveness of SfM methods on datasets with varying levels of sparsity.
The best results of COLMAP and HLOC can be found in Table~\ref{table:colmap_result}.
Furthermore, we employ LocalRF~\cite{meuleman2023localrf} to ensure we can estimate all camera poses.
We have made modifications to the code provided in the LocalRF paper to restrict the sampling of rays solely to valid color pixels.
Table~\ref{table:colmap_result} indicates that COLMAP exhibits a substantially higher speed than HLOC, with a speedup factor of 8 and 26.5 for $n_{frame}=\{310,1392\}$, respectively. 
However, COLMAP only performs well in the case of a dense view with $n_{frame}=2784$, whereas HLOC demonstrates better effectiveness in other scenarios with $n_{frame}=\{310,1392\}$.
Although LocalRF~\cite{meuleman2023localrf} is capable of estimating all camera poses, its extensive running time presents a drawback.
Consequently, LocalRF is only suitable for application in sparse views, where the number of images that needs to be determined is limited.

\begin{table}[tb]
    \centering
    \caption{
        Results of camera pose estimation.
        We conducted experiments with different settings and recorded the number of frames, denoted as $n_{pose}$, for which camera poses were successfully estimated.
        The highest $n_{pose}$ (excluding LocalRF~\cite{meuleman2023localrf}) is shown in \textcolor{red}{\textbf{bold}}, and the second best is \textcolor{blue}{\underline{underscored}} for each $n_{frame}$ dataset.
        LocalRF is capable of estimating all camera poses, albeit with a trade-off in terms of running time.
        Similarly, HLOC suffers from long running times.
        Consequently, we did not use HLOC and LocalRF for $n_{frame}=2784$.
    }
    \label{table:colmap_result}
    \begin{tabularx}{\textwidth}{|c|>{\centering\arraybackslash}X|>{\centering\arraybackslash}X|>{\centering\arraybackslash}X|}
    \toprule
    \textbf{$n_{frame}$}                    & \textbf{Method}                   & \textbf{$n_{pose}$}                   & \textbf{Time (hh:mm:ss)} \\ \midrule
    \multirow{3}{*}{310}                    & COLMAP                            & \textcolor{red}{\textbf{62}}          & 00:08:26               \\ 
                                            & HLOC                              & \textcolor{blue}{\underline{43}}      & 01:08:39               \\ 
                                            & LocalRF                           & 310                                   & 23:20:52               \\ \midrule
    \multirow{3}{*}{1392}                   & COLMAP                            & \textcolor{blue}{\underline{376}}     & 00:49:29               \\ 
                                            & HLOC                              & \textcolor{red}{\textbf{647}}         & 21:51:34               \\ 
                                            & LocalRF                           & 1392                                  & 75:10:54               \\ \midrule
    2784                                    & COLMAP                            & \textcolor{red}{\textbf{2667}}        & 04:32:53               \\
    \bottomrule
    \end{tabularx}
\end{table}
%
%

\subsection{Qualitative and Quantitative Rendering Results}
Based on the results of the previous experiment, 
we created individual image sets for five possible $n_{pose}$ values: $\{62, 310, 647, 1392, 2667\}$. 
We then trained NeRFs-based models on each of these sets. 
For training, we took 90\% of $n_{pose}$, comprising equally spaced images, as the training set. 
The remaining images were allocated to the evaluation set.
To avoid memory limitations when working with a large number of 4K images, we down-scaled the input images by a factor of four. 
We trained the NerfAcc~\cite{li2023nerfacc} and Instant-NGP~\cite{mueller2022instant} models using the default hyperparameter settings provided by the Nerfstudio library~\cite{nerfstudio}. 
We observed that training with a sampling mask increased the processing time 
because of the underlying Nerfstudio implementation.
Therefore, we adjusted the number of training iterations to be $\{3000, 15000, 30000, 50000, 100000\}$, corresponding to the number of images in each $n_{pose}$ set, respectively.

For qualitative evaluation, Fig.~\ref{fig:woMask_problem} shows the render results obtained for a select number of images in the evaluation set.
As explained in Section~\ref{subsec:method_NeRF}, training without a mask resulted in black patterns, while training with a mask improved the image quality and expanded the field of view.
However, it is important to note that training with a mask currently presents a trade-off between image quality and processing time.
Overall, the rendered frames from Instant-NGP~\cite{mueller2022instant} exhibited higher quality and less blurriness compared to NerfAcc~\cite{li2023nerfacc}.
Furthermore, among the different values of $n_{pose}$, the set with $n_{pose}=2667$ (dense views) yielded the highest quality rendered frames compared to the other sets. 
This finding is consistent with the notion that a NeRFs-based method generally performs better 
when this method is applied to dense and overlapping views.

\begin{figure}[tb]
    \centering
    \includegraphics[width=1\textwidth]{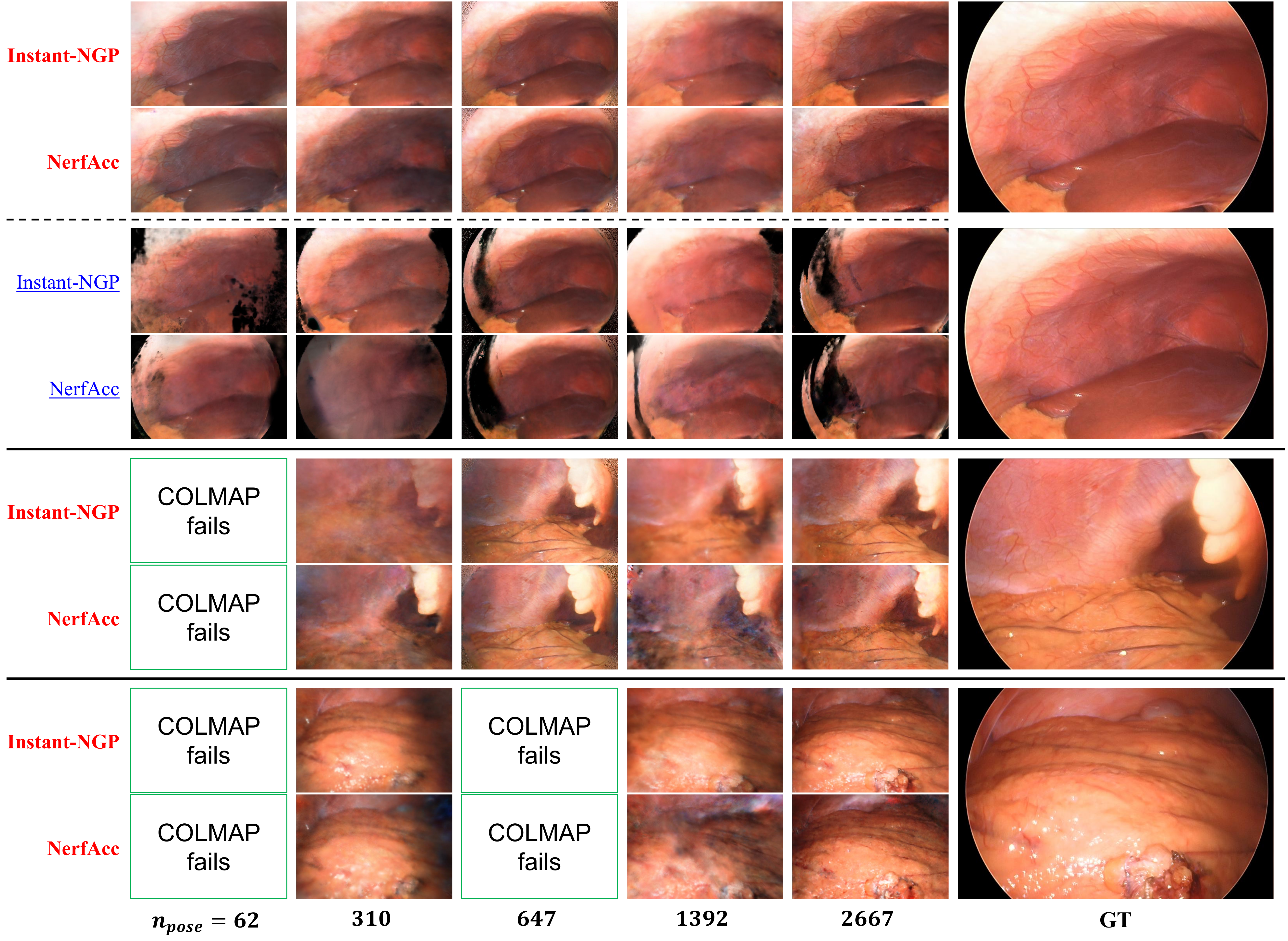}
    \caption{
    Novel view synthesis for a select number of unseen video frames in the evaluation set.
    Training with a mask is shown in \textcolor{red}{\textbf{bold}}, whereas training without a mask is shown as \textcolor{blue}{\underline{underscored}}.
    Black patterns occur when training without a mask.
    The green box indicates a failure of COLMAP to estimate the camera pose of the video frame in the $n_{pose}$ set (rendering a video frame is not possible without the correct camera perspective).
    The frames generated by Instant-NGP~\cite{mueller2022instant} are better in subjective quality and less blurry than those produced by NerfAcc~\cite{li2023nerfacc}.
    }
    \label{fig:NeRF_result}
\end{figure}

To evaluate the objective quality of the synthesized novel views, we measure the PSNR, SSIM, and LPIPS~\cite{zhang2018unreasonable} scores, which are conventional image quality metrics~\cite{mildenhall2020nerf,wang2022neural}, between the synthesized views and the corresponding ground truth video frames.
The results obtained for training with a mask are presented in Table~\ref{table:nerf_result}.
We can observe that these results are in line with our qualitative findings, suggesting that Instant-NGP~\cite{mueller2022instant} outperforms NerfAcc~\cite{li2023nerfacc}.

\begin{table}[htb]
    \centering
    \caption{
        Quantitative results obtained for video frames in the evaluation set.
        We report the average PSNR/SSIM scores (higher is better) and LPIPS~\cite{zhang2018unreasonable} scores (lower is better). 
        The highest score for each $n_{pose}$ evaluation set is indicated in \textcolor{red}{\textbf{bold}}. 
        Overall, the Instant-NGP~\cite{mueller2022instant} model outperforms the NerfAcc~\cite{li2023nerfacc} model across all $n_{pose}$ sets.
    }
    \label{table:nerf_result}
    
    \begin{tabularx}{\textwidth}{|c|*4{>{\centering\arraybackslash}X|}}
    \toprule
    \textbf{$n_{pose}$}     & \textbf{Method}   & \textbf{PSNR~$\uparrow$}                      & \textbf{SSIM~$\uparrow$}                      & \textbf{LPIPS~$\downarrow$}                   \\ \midrule
    \multirow{2}{*}{62}     & NerfAcc           & 12.760 $\pm$ 0.424                            & \textcolor{red}{\textbf{0.764 $\pm$ 0.018}}   & \textcolor{red}{\textbf{0.448 $\pm$ 0.020}}   \\
                            & Instant-NGP       & \textcolor{red}{\textbf{12.769 $\pm$ 0.231}}  & 0.763 $\pm$ 0.008                             & 0.470 $\pm$ 0.028                             \\ \midrule
    \multirow{2}{*}{310}    & NerfAcc           & 12.520 $\pm$ 2.439                            & 0.674 $\pm$ 0.084                             & 0.619 $\pm$ 0.097                             \\
                            & Instant-NGP       & \textcolor{red}{\textbf{13.558 $\pm$ 1.620}}  & \textcolor{red}{\textbf{0.714 $\pm$ 0.051}}   & \textcolor{red}{\textbf{0.609 $\pm$ 0.099}}   \\ \midrule
    \multirow{2}{*}{647}    & NerfAcc           & 13.817 $\pm$ 1.447                            & 0.730 $\pm$ 0.044                             & 0.510 $\pm$ 0.084                             \\
                            & Instant-NGP       & \textcolor{red}{\textbf{14.136 $\pm$ 1.350}}  & \textcolor{red}{\textbf{0.744 $\pm$ 0.034}}   & \textcolor{red}{\textbf{0.472 $\pm$ 0.053}}   \\ \midrule
    \multirow{2}{*}{1392}   & NerfAcc           & 13.602 $\pm$ 1.598                            & 0.704 $\pm$ 0.044                             & 0.548 $\pm$ 0.055                             \\
                            & Instant-NGP       & \textcolor{red}{\textbf{14.694 $\pm$ 1.244}}  & \textcolor{red}{\textbf{0.746 $\pm$ 0.025}}   & \textcolor{red}{\textbf{0.538 $\pm$ 0.051}}   \\ \midrule
    \multirow{2}{*}{2667}   & NerfAcc           & 14.226 $\pm$ 1.526                            & 0.713 $\pm$ 0.039                             & 0.529 $\pm$ 0.058                             \\ 
                            & Instant-NGP       & \textcolor{red}{\textbf{14.512 $\pm$ 1.549}}  & \textcolor{red}{\textbf{0.757 $\pm$ 0.025}}   & \textcolor{red}{\textbf{0.450 $\pm$ 0.050}}   \\ \bottomrule                                     
    \end{tabularx}
\end{table}

\section{Conclusions}
 In this paper, we introduced a workflow for applying state-of-the-art NeRFs-based methods to laparoscopic surgical videos in order to render a complete abdominal environment, identifying and overcoming several challenges. 
 
A first challenge we encountered is the presence of artifacts in recorded video frames (e.g., blurriness caused by laparoscope movement), leading to failures in camera pose estimation using COLMAP.
To mitigate these failures, we utilized LocalRF~\cite{meuleman2023localrf}, which enabled us to estimate all camera poses but required a substantial amount of computational time (e.g., 75 hours for 1392 images).
%
%
%

A second challenge we faced was determining the appropriate frame resolution and the number of extracted frames ($n_{frame}$). 
Opting for frames with a higher resolution, such as 4K or even 8K~\cite{yamashita2016ultra}, resulted in rendered images of a higher quality but also increased training time.
This is due to the need to randomly sample rays (pixels) in each training iteration to ensure full pixel coverage.
Also, larger $n_{pose}$ values substantially increased the training time. 
Consequently, a trade-off exists between training time and rendered image quality.
%
%

A third challenge we came across is the limited view provided by a laparoscope, resulting in recorded videos that typically contain black outer areas. 
To address this challenge, we employed sampling with a mask as a workaround.
As shown by our experimental results, the proposed one-for-all mask approach enhanced the quality of the rendered outcomes and broadened the perspective.
%
%

A fourth challenge is related to the camera trajectory followed, which is ---in the context of laparoscopy surgery--- usually free in nature, rather than being a regular forward-facing trajectory that captures extensive overlapping views. 
Such an elongated and unconstrained trajectory poses difficulties for NeRFs-based models to fully render an abdominal scene.

%
In future work, we will keep focussing on overcoming the primary challenge of the limited surgical view provided by a laparoscope. 
To that end, we aim at combining NeRFs with diffusion models~\cite{dhariwal2021diffusion,ho2020denoising,sohl2015deep}, such as DiffRF~\cite{muller2023diffrf}, NerfDiff~\cite{gu2023nerfdiff}, and RealFusion~\cite{melaskyriazi2023realfusion}.
Additionally, we plan to employ NeRFs to predict intraoperative changes during laparoscopic surgery, for instance addressing the issue of organ deformation between preoperative and intraoperative states.

%
%
\bibliographystyle{splncs04}
\bibliography{ref}
%




\end{document}